\documentclass[10pt,conference]{IEEEtran}
\IEEEoverridecommandlockouts

\usepackage[T1]{fontenc}
\usepackage[utf8]{inputenc}
\usepackage{microtype}
\usepackage{graphicx}
\usepackage{listings}
\usepackage{xurl} 
\usepackage{hyperref}
\usepackage{booktabs}
\usepackage{tabularx}
\usepackage{array}
\usepackage{multirow}
\usepackage{xcolor}
\usepackage{amsmath}
\usepackage{xcolor}
\usepackage{float}
\usepackage{graphicx}

\usepackage[T1]{fontenc}
\usepackage{inconsolata} 
\usepackage{xcolor}
\usepackage{listings}

\definecolor{codebg}{RGB}{248,248,248}
\definecolor{cframe}{RGB}{220,220,220}
\definecolor{ckey}{RGB}{0,63,135}
\definecolor{cstring}{RGB}{163,21,21}
\definecolor{ccomment}{RGB}{106,153,85}
\definecolor{cnumber}{RGB}{128,128,128}

\lstdefinestyle{pytheme}{
  language=Python,
  backgroundcolor=\color{codebg},
  basicstyle=\ttfamily\footnotesize,
  keywordstyle=\color{ckey}\bfseries,
  stringstyle=\color{cstring},
  commentstyle=\color{ccomment}\itshape,
  numberstyle=\tiny\color{cnumber},
  numbers=left,
  stepnumber=1,
  numbersep=8pt,
  frame=single,
  rulecolor=\color{cframe},
  frameround=tttt,
  breaklines=true,
  breakatwhitespace=false,
  showstringspaces=false,
  columns=fullflexible,
  keepspaces=true,
  tabsize=4,
  numbers=none,
}

\title{\includegraphics[height=1.1em]{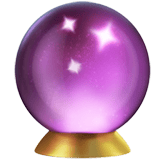} PrismSSL: One Interface, Many Modalities; A Single-Interface Library for Multimodal Self-Supervised Learning}

\makeatletter
\newcommand{\linebreakand}{%
  \end{@IEEEauthorhalign}
  \hfill\mbox{}\par
  \mbox{}\hfill\begin{@IEEEauthorhalign}
}
\makeatother

\author{\IEEEauthorblockN{1\textsuperscript{st} Melika Shirian}
\IEEEauthorblockA{\textit{dept. Computer Engineering} \\
\textit{University of Isfahan}\\
Isfahan, Iran}
\and

\IEEEauthorblockN{1\textsuperscript{st} Kianoosh Vadaei}
\IEEEauthorblockA{\textit{dept. Computer Engineering} \\
\textit{University of Isfahan}\\
Isfahan, Iran }
\linebreakand 

\IEEEauthorblockN{2\textsuperscript{nd} Kian Majlessi}
\IEEEauthorblockA{\textit{dept. Computer Engineering} \\
\textit{University of Isfahan}\\
Isfahan, Iran }
\and

\IEEEauthorblockN{2\textsuperscript{nd} Audrina Ebrahimi}
\IEEEauthorblockA{\textit{dept. Electrical and Computer Engineering} \\
\textit{University of Texas at Dallas}\\
Richardson, TX, USA }

\linebreakand 

\IEEEauthorblockN{3\textsuperscript{rd} Peyman Adibi}
\IEEEauthorblockA{\textit{dept. Artificial Intelligence} \\
\textit{University of Isfahan}\\
Isfahan, Iran }
\and
\IEEEauthorblockN{3\textsuperscript{rd} Hossein Karshenas}
\IEEEauthorblockA{\textit{dept. Artificial Intelligence} \\
\textit{University of Isfahan}\\
Isfahan, Iran}
}

\begin{document}
\maketitle

\begin{abstract}
We present \textbf{PrismSSL}, a Python library that unifies state-of-the-art self-supervised learning (SSL) methods across \emph{audio}, \emph{vision}, \emph{graphs}, and \emph{cross-modal} settings in a single, modular codebase. The goal of the demo is to show how researchers and practitioners can: (i) install, configure, and run pretext training with a few lines of code; (ii) reproduce compact benchmarks; and (iii) extend the framework with new modalities or methods through clean trainer and dataset abstractions. PrismSSL is packaged on PyPI, released under the MIT license, integrates tightly with HuggingFace Transformers~\cite{wolf-etal-2020-transformers}, and provides quality-of-life features such as distributed training in PyTorch~\cite{pytorch}, Optuna-based hyperparameter search~\cite{optuna}, LoRA fine-tuning for Transformer backbones~\cite{lora}, animated embedding visualizations for sanity checks, Weights \& Biases logging~\cite{wandb}, and colorful, structured terminal logs for improved usability and clarity. In addition, PrismSSL offers a graphical dashboard---built with Flask~\cite{flask} and standard web technologies---that enables users to configure and launch training pipelines with minimal coding. The artifact (code and data recipes) is publicly available and reproducible.\footnote{Docs: \url{https://prismaticlab.github.io/PrismSSL}\;|\;Code: \url{https://github.com/PrismaticLab/PrismSSL}.}
\end{abstract}

\section{Introduction}
Self-supervised learning (SSL) learns useful representations by solving pretext tasks on raw data, removing the need for human labels. In vision, contrastive and non-contrastive objectives such as SimCLR, MoCo, BYOL, and DINO have shown strong transfer; masked image modeling (MAE) further narrows the gap to supervised pretraining~\cite{simclr,moco,byol,dino,mae}. In audio, predictive objectives like wav2vec~2.0 and HuBERT learn from waveforms without transcripts~\cite{wav2vec2,hubert}. On graphs and across modalities, methods such as GraphCL (graphs) and CLIP (image–text) broaden SSL beyond single domains~\cite{graphcl,clip}.

\begin{figure}[t]
    \centering
    \includegraphics[width=0.46\textwidth]{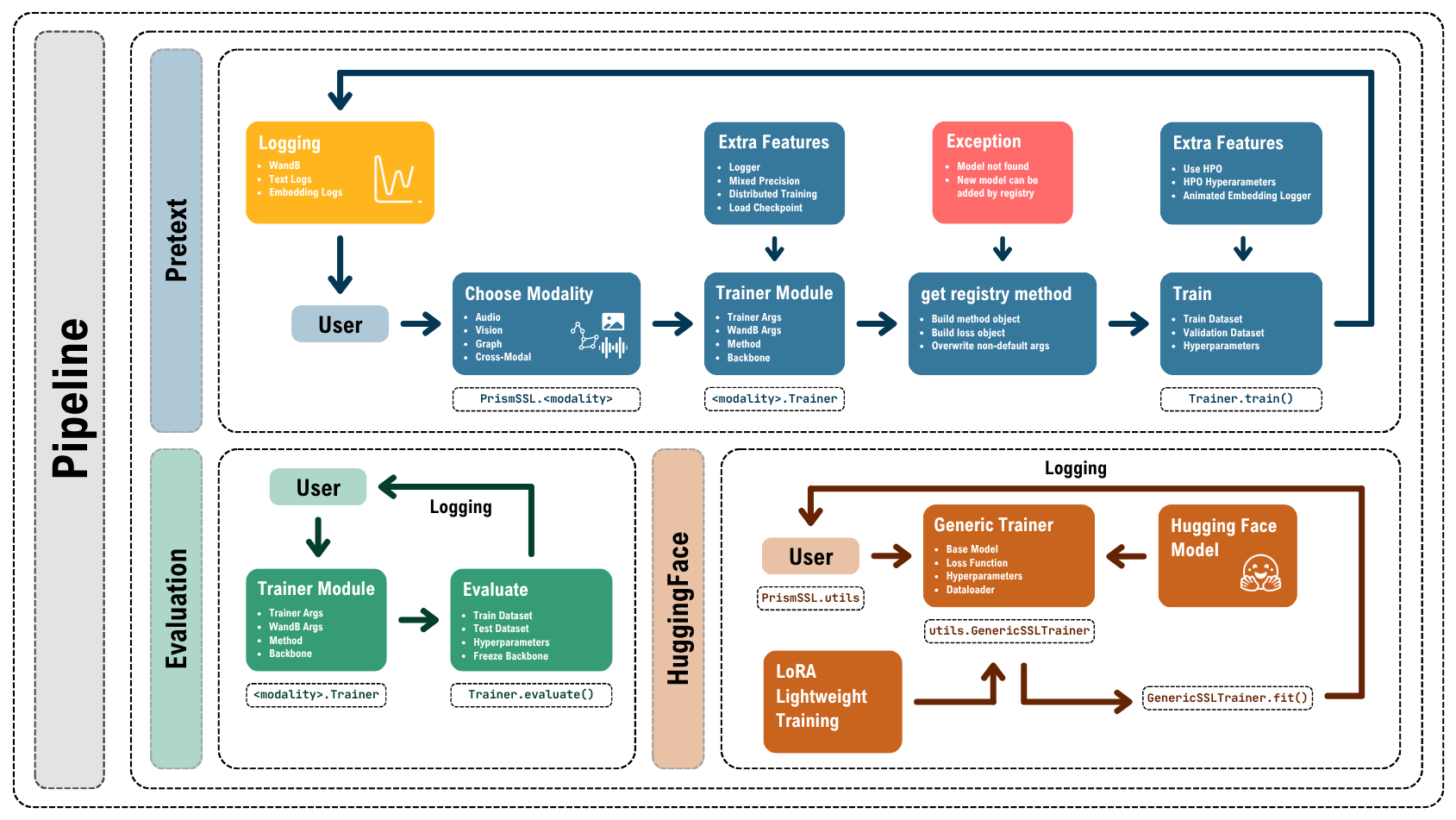}
    \caption{PrismSSL pipeline overview.}
    \label{fig:pipeline}
\end{figure}

\textbf{Challenges.} Despite rapid progress, research practice remains fragmented: implementations are spread across domain-specific repositories with incompatible data loaders, augmentations, and training loops, which complicates fair comparison and reproducibility across modalities. This fragmentation slows iteration and obstructs methodological synthesis.

\textbf{Design goals.} PrismSSL targets four practical goals: 
{G1 (Unified UX)}---a single mental model for installing, configuring, training, and evaluating SSL across modalities; 
G2 (Reproducibility)---deterministic configs, logged runs, and \emph{exportable} checkpoints with compact benchmark suites; 
G3 (Extensibility)---clear interfaces for adding methods, augmentations, loss functions, and backbones; and 
G4 (Practicality)---first-class support for distributed training, resource-aware batching, LoRA fine-tuning, hyperparameter optimization (HPO), and an optional graphical configuration UI.

\textbf{At a glance.} \textbf{PrismSSL} provides a single, pragmatic interface to run SSL across \emph{audio}, \emph{vision}, \emph{graphs}, and \emph{cross-modal} data. Users select a modality and method via a registry, plug in either a default backbone from the library or a user-defined PyTorch backbone, and launch pretext or evaluation with the same API; see Fig.~\ref{fig:pipeline}. Quality-of-life features---Optuna HPO~\cite{optuna}, LoRA fine-tuning for integrated Transformer models~\cite{lora}, distributed training in PyTorch~\cite{pytorch}, and experiment tracking via W\&B~\cite{wandb}---are integrated but optional; HuggingFace ecosystem compatibility is provided out of the box~\cite{wolf-etal-2020-transformers}.

\textbf{Supported modalities.} Table~\ref{tab:intro-modalities} summarizes the modalities currently covered by PrismSSL, including representative SSL methods and typical default or user-provided PyTorch backbones. Cross-modal coverage spans audio–image–text models such as CLAP, AudioCLIP, Wav2CLIP, ALBEF, SimVLM, SLIP, UNITER, and VSE++~\cite{clap,audioclip,wav2clip,albef,simvlm,slip,uniter,vsepp}.

\textbf{Contributions.}
\begin{itemize}
    \item \textbf{C1}:\; A unified trainer and registry that decouple \emph{method}, \emph{data}, and \emph{runtime} layers.
    \item \textbf{C2}:\; A reproducible artifact with small yet representative benchmarks and a step-by-step usage scenario.
    \item \textbf{C3}:\; Plug-ins for HuggingFace backbones, Optuna HPO, DDP, and LoRA.
    \item \textbf{C4}:\; An extensibility recipe (\emph{add-a-method}) enabling new SSL objectives with minimal boilerplate.
\end{itemize}

\begin{table*}[t]
\centering
\caption{Modalities supported by PrismSSL with representative SSL methods and example default backbones.}
\label{tab:intro-modalities}
\footnotesize
\setlength{\tabcolsep}{2pt}
\renewcommand{\arraystretch}{1.05}
{\setlength{\emergencystretch}{2em}%
\begin{tabular*}{\textwidth}{@{} >{\raggedright\arraybackslash}p{0.15\textwidth} @{\hspace{1pt}} >{\raggedright\arraybackslash}p{0.66\textwidth} @{\hspace{1pt}} >{\raggedright\arraybackslash}p{0.20\textwidth} @{}}
\toprule
\textbf{Modality} & \textbf{SSL methods (examples)} & \textbf{Default backbones (examples)} \\
\midrule
Audio & wav2vec2, COLA, SpeechSimCLR, EAT & Transformer, 1D CNN \\
Vision & MAE, SimCLR, BYOL, DINO, Barlow Twins, MoCo v2--v3, SimSiam, SwAV & ViT-B/16, Resnet18 \\
Cross-Modal & CLAP, AudioCLIP, Wav2CLIP, CLIP, ALBEF, SLIP, SimVLM, UNITER, VSE & ViT-B/32, BERT \\
Graphs & GraphCL & GIN, GCN \\
\bottomrule
\end{tabular*}}
\end{table*}

\section{PRISM Architecture}
\begin{figure}[t]
    \centering
    \includegraphics[width=0.33\textwidth]{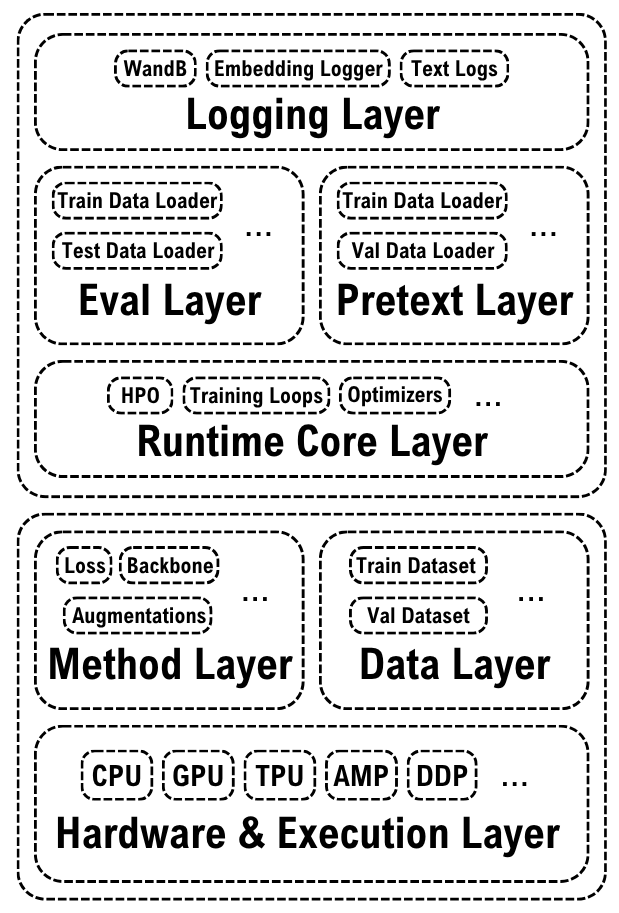}
    \caption{PrismSSL layered architecture overview.}
    \label{fig:arch}
\end{figure}

PrismSSL is implemented in Python on top of PyTorch~\cite{pytorch}. As shown in Fig.~\ref{fig:arch}, a modality-specific \texttt{Trainer} coordinates (i) data modules (datasets, transforms, \texttt{collate}); (ii) method components (including loss/objective modules, main implementation modules, and projection heads); (iii) backbone encoders; and (iv) runtime utilities (logging, checkpointing, HPO, DDP).

\begin{figure}[!htbp]
    \centering
    \includegraphics[width=0.46\textwidth]{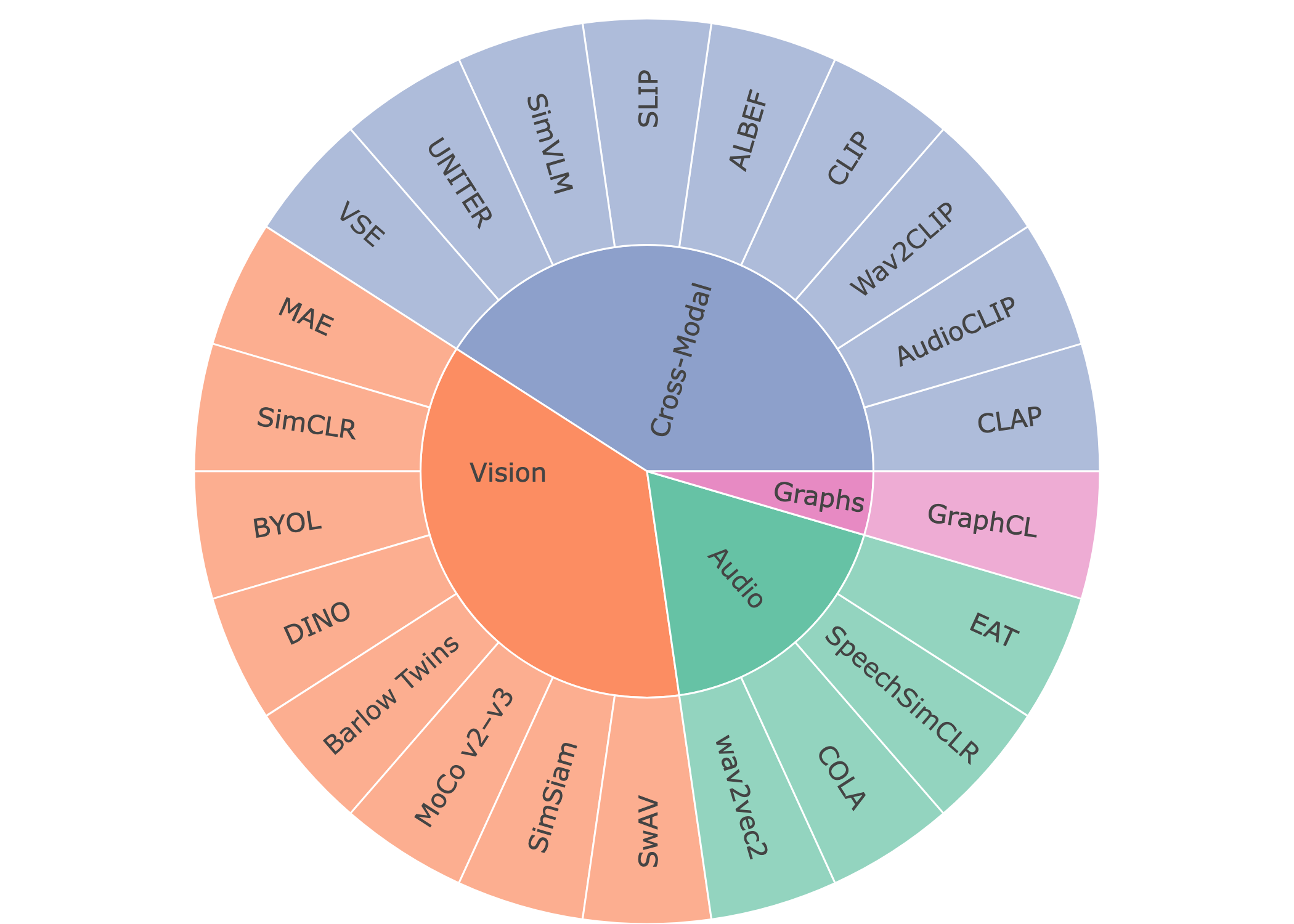}
    \caption{Proportional overview of supported PrismSSL modalities and methods.}
    \label{fig:modalities}
\end{figure}

\subsection{Supported Methods (v0.2.1)}
PrismSSL currently implements the following SSL families per modality. Each method is a modular \emph{config\,+\,loss} component selectable by a string key (see \S\ref{sec:usage}).
\begin{itemize}
  \item \textbf{Audio}: predictive/contrastive objectives, including \emph{wav2vec~2.0}~\cite{wav2vec2}, \emph{HuBERT}~\cite{hubert}, SpeechSimCLR, COLA, and EAT.
  \item \textbf{Vision}: masked modeling and instance-learning variants---MAE~\cite{mae}, SimCLR~\cite{simclr}, BYOL~\cite{byol}, DINO~\cite{dino}, Barlow Twins, MoCo v2/v3~\cite{moco}, SimSiam, and SwAV.
  \item \textbf{Graphs}: contrastive graph pretraining---GraphCL~\cite{graphcl}.
  \item \textbf{Cross-modal}: audio–image–text alignment---CLAP~\cite{clap}, AudioCLIP~\cite{audioclip}, Wav2CLIP~\cite{wav2clip}, CLIP~\cite{clip}, ALBEF~\cite{albef}, SLIP~\cite{slip}, SimVLM~\cite{simvlm}, UNITER~\cite{uniter}, and VSE++~\cite{vsepp}.
\end{itemize}

\subsection{Modalities}
PrismSSL provides thin modality adapters over a shared trainer:
\begin{itemize}
  \item \textbf{Audio}-waveform or spectrogram inputs; modality-specific augmentations and a padding-aware \texttt{collate}.
  \item \textbf{Vision}-image/video inputs; standard crop/flip/color-jitter pipelines; multi-view support for contrastive methods.
  \item \textbf{Cross-modal}-audio–image–text alignment pipelines (e.g., CLAP/CLIP family).
  \item \textbf{Graphs}-molecular/social graphs with mini-batching in the style of PyTorch Geometric~\cite{pyg}.
\end{itemize}

\subsection{Models}
Methods define the objective (contrastive, masked modeling, bootstrap), while backbones are pluggable encoders. Defaults are provided per modality (e.g., ViT/ResNet18 for vision, transformer encoder/1D CNN for audio, GIN/GCN for graphs), and cross-modal setups (e.g., CLIP for image-text) are supported. Defaults are selected according to their original papers, and users may attach custom PyTorch modules.

\subsection{Generic Trainer Module}\label{sec:hf}
This optional module enables standalone HuggingFace interoperability~\cite{wolf-etal-2020-transformers}. Users can import HF models, define custom loss functions (with automatic parameter detection via Python function \emph{signatures}), and either train from scratch or continue from existing weights through a purpose-built API. Low-rank adapters (LoRA)~\cite{lora} can be attached for efficient fine-tuning; \emph{rank}, target modules, dropout, $\alpha$, and \texttt{task\_type} are configurable.

\subsection{Trainer Module}
The trainer is implemented as two main components (training and evaluation) exposing:
\begin{enumerate}
    \item \textbf{Library APIs} (\S\ref{sec:usage}) for programmatic use.
    \item An optional web \textbf{UI} (Flask~\cite{flask} with HTML/CSS/JS) to:
    \begin{itemize}
      \item set training configs (method, backbone, batch size, optimizer),
      \item attach a PyTorch \texttt{Dataset} or load a recipe,
      \item import/export JSON configs for repeatability,
      \item start pretext/evaluation (GUI dashboard currently supports pretext only) with live terminal logs.

    \end{itemize}
\end{enumerate}

\subsection{Registry}
Methods are selected by a string key (e.g., \texttt{method=\string"mae\string"}); adding a new method registers a factory plus a small config schema.

\subsection{Hyperparameter Optimization (HPO)}
Optuna~\cite{optuna} trials can be enabled via a flag. Typical search spaces include \textit{learning\_rate}, \textit{weight\_decay}, \textit{batch\_size}; \textit{n\_trials} and \textit{tuning\_epochs} are configurable, with pruning for early stopping on intermediate metrics.

\subsection{Distributed Training}
Multi-GPU/host training uses PyTorch DataParallel (DD)~\cite{pytorch}, with optional AMP mixed precision.

\subsection{Weights \& Biases (W\&B)}
All runs log to W\&B~\cite{wandb}:
\begin{enumerate}
  \item live \emph{epoch}-level loss plots,
  \item live \emph{batch}-level loss plots,
  \item system metrics (GPU/CPU/memory),
  \item animated embedding logging (sanity checks),
  \item artifact saving (checkpoints \& configs),
  \item full config logging for exact reproducibility.
\end{enumerate}

\subsection{Terminal Logging}
A structured, colorized logger (built on the Python \texttt{logging} library) mirrors W\&B metrics locally, printing initial summaries and using \texttt{tqdm} for progress, with final run summaries logged to the terminal.

\subsection{Future Directions}
Broaden modality coverage (e.g., AV-HuBERT), expand backbone families, and add continuous benchmarking with auto-dataset recipes.

\section{Tool Usage Scenario}
\label{sec:usage}
\noindent
\textbf{PrismSSL} is a modular Python toolbox for self-supervised learning, installable via \texttt{pip install prism-ssl} and compatible with any OS running Python~3.8+ (CUDA is recommended). Once installed, users provide data via a standard PyTorch \texttt{Dataset}~\cite{pytorch}. PrismSSL supports two complementary workflows: a \emph{Python-based} API for full control, and an \emph{optional UI} for quick prototyping. Experiments can log to Weights\&Biases for tracking and sharing~\cite{wandb}; interoperability with HuggingFace Transformers is provided through a separate API for model training~\cite{wolf-etal-2020-transformers}.

\subsection{Python-Based Usage}
Users import PrismSSL modules into scripts or notebooks to configure \emph{pretext} training and \emph{evaluation}. The example below runs audio pretraining with Wav2Vec2 and then evaluates via a linear probe. (See \S\ref{sec:usage} for more end-to-end flows.)

\vspace{3cm}
\renewcommand{\lstlistingname}{Listing}
\begin{lstlisting}[
  style=pytheme,
  language=Python,
  caption={Minimal audio pretext + evaluation with the modality-specific Trainer.},
  label={lst:minimal-audio-trainer},
  captionpos=b
]
from PrismSSL.audio.Trainer import Trainer

trainer = Trainer(
    method='wav2vec2',
    backbone=None,               # or a custom PyTorch module
    save_dir='./',
    wandb_project='wav2vec2-pretext',
    wandb_mode='online',         # requires W&B login
    use_data_parallel=True,
    checkpoint_interval=10,
    verbose=True,
    reload_checkpoint=False,
    mixed_precision_training=False,
)

trainer.train(
    train_dataset=train_dataset,
    val_dataset=val_dataset,    # optional validation dataset
    batch_size=16,
    epochs=100,
    learning_rate=1e-4,
    weight_decay=1e-2,
    optimizer='adamw',  # one of ['adam', 'adamw', 'sgd']
    use_hpo=True, n_trials=20, tuning_epochs=5,  # Optuna HPO
    use_embedding_logger=True, logger_loader=logger_loader
)

trainer.evaluate(
    train_dataset=train_dataset, test_dataset=test_dataset,
    num_classes=39, batch_size=64, lr=1e-3, epochs=10,
    freeze_backbone=True
)
\end{lstlisting}

\medskip
\renewcommand{\lstlistingname}{Listing}
\begin{lstlisting}[
  style=pytheme,
  language=Python,
  caption={Generic SSL trainer with a HuggingFace backbone (illustrative).},
  label={lst:minimal-generictrainer-hf},
  captionpos=b
]
from transformers import BertForPreTraining, AutoTokenizer  # 
model = BertForPreTraining.from_pretrained("bert-base-uncased")
tokenizer = AutoTokenizer.from_pretrained("bert-base-uncased")

trainer = GenericSSLTrainer(
    model=model,
    loss_fn=bert_loss_fn,
    dataloader=dataloader,
    optimizer_ctor=optimizer,
    epochs=10,
    # Optional LoRA configuration (if enabled)
    use_lora=False, r=8, lora_alpha=32,
    target_modules=["query", "key", "value"],
    lora_dropout=0.1, bias="none",
    task_type="FEATURE_EXTRACTION",
)
trainer.fit()
\end{lstlisting}

\subsection{UI-Based Usage}
\begin{figure}[H]
    \centering
    \includegraphics[width=0.42\textwidth]{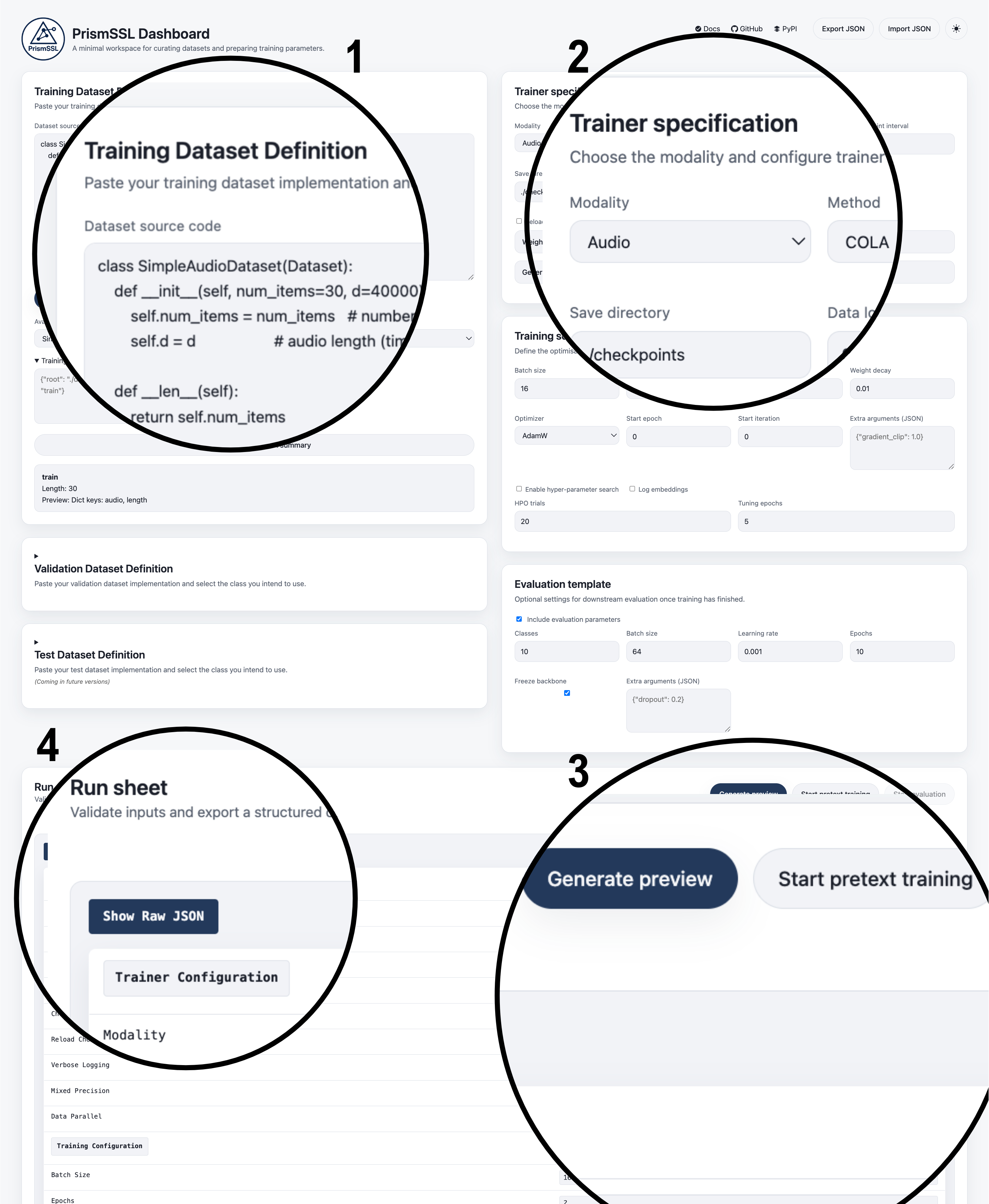}
    \caption{PrismSSL dashboard for low-/no-code experimentation. The UI lets users (1) paste a PyTorch Dataset for train/val (test support coming in future versions), (2) select modality and SSL method in the Trainer specification and set core hyperparameters (batch size, optimizer, LR/WD, HPO, paths), (3) optionally define an evaluation template (classes, epochs, freeze-backbone), (4) review the structured run sheet and export/import JSON, and (5) launch runs via ‘Generate preview’ or ‘Start pretext training.’ All choices compile into a reproducible configuration.}
    \label{fig:ui-fig}
\end{figure}

Alternatively, users can launch the integrated user interface for a low-code workflow. The UI lets users supply their \texttt{Dataset} code, choose the modality/method, set core hyperparameters, and start runs with minimal scripting. Results are logged just as in the Python-based workflow for reproducibility and sharing. This mode is ideal for quick ablations and comparative runs before switching to scripted pipelines, with the overall process illustrated in Fig.~\ref{fig:ui-fig}.

\section{Case Study}
We verified every implemented model in \textbf{PrismSSL} with lightweight sanity checks; here we report one representative example using \textbf{Wav2CLIP}. We assembled a binary \emph{cats} vs.\ \emph{dogs} image set and, for each class, authored seven short prompts (e.g., \emph{``a beautiful photo of a dog''}). Each sentence was synthesized to speech via \texttt{Gemini~2~Flash~TTS} with randomized speed, pitch, and speaker identity to induce acoustic diversity. The audio clips were encoded and mean-pooled per class to form a prototype matrix $(C,D)$; images were encoded into a batch matrix $(B,D)$.

Class scores were obtained by a dot product between $(B,D)$ and $(C,D)^{\mathsf T}$, followed by a log-softmax over classes to yield zero-shot probabilities. This procedure tests both the data pathway and the alignment objective without any fine-tuning.

\begin{figure}[!htbp]
    \centering
    \includegraphics[width=0.37\textwidth]{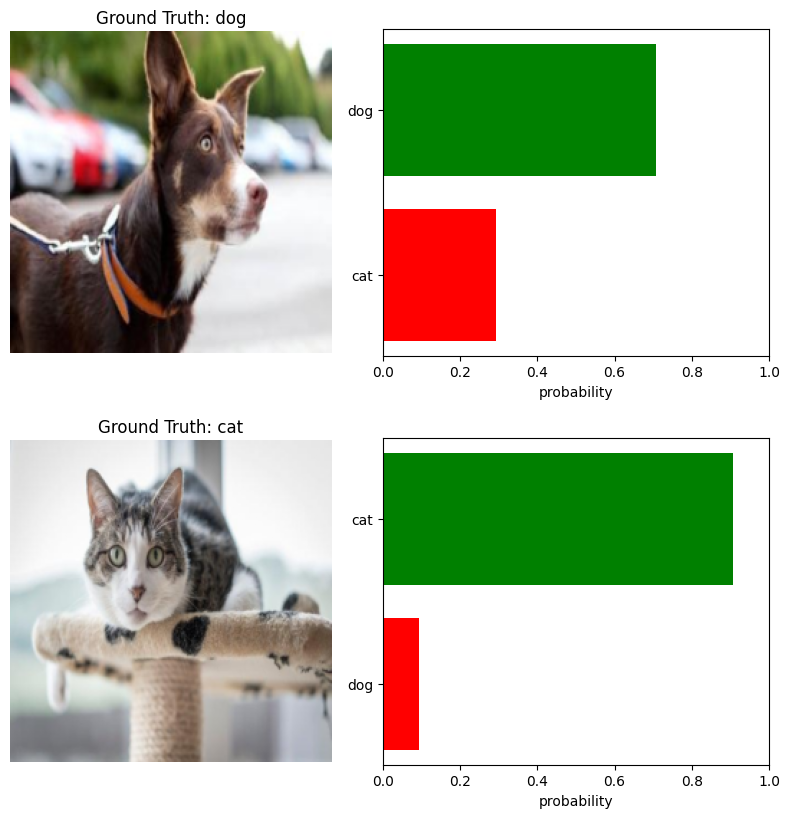}
    \caption{Zero-shot Wav2CLIP probabilities on the cat–dog set.}
    \label{fig:cat_dog_prob}

\end{figure}

\begin{figure}[!htbp]
    \centering
    \includegraphics[width=0.37\textwidth]{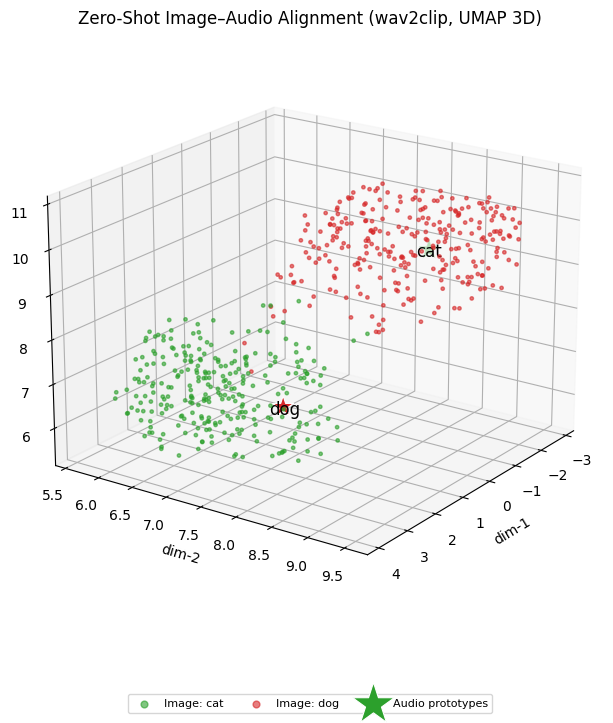}
    \caption{Wav2CLIP image–audio alignment (UMAP 3D).}
    \label{fig:wav2clip_zero_shot}
\end{figure}

Fig.~\ref{fig:cat_dog_prob} and Fig.~\ref{fig:wav2clip_zero_shot} illustrate that audio prototypes align closely with their visual counterparts, producing clear class separation in latent space and confident zero-shot predictions. While this example highlights Wav2CLIP, the same prototype-based evaluation was extended to other cross-modal models, whereas single-modality methods were validated through their respective modality-specific checks.

\section{Limitations and Future Work}
We currently ship representative methods per modality. Future work includes broader backbone coverage (e.g., Conformer/HTSAT for audio, ViT variants for vision, diverse GNNs for graphs), richer cross-modal recipes (e.g., AV-HuBERT), expanded unit tests, and continuous benchmarking pipelines.

\section{Conclusion}
PrismSSL lowers the entry barrier to multi-domain self-supervision by standardizing trainers, data collation, and runtime features across modalities. We invite the community to use, extend, and contribute methods and datasets.

\section*{Acknowledgments}
We thank all contributors and users who tested early prototypes and reported issues.

\end{document}